\documentclass[10pt,journal,final]{IEEEtran}
\IEEEoverridecommandlockouts

\usepackage{amsfonts,amssymb}
\usepackage{ifpdf}
\usepackage{cite}
\usepackage{multirow}
\usepackage[hyphens]{url}
\usepackage{stfloats}
\usepackage{bm}
\usepackage{color,xcolor}
\usepackage{subfigure}
\usepackage{caption}
\ifCLASSINFOpdf
   \usepackage[pdftex]{graphicx}
   \graphicspath{{../pdf/}{../jpeg/}}
   \DeclareGraphicsExtensions{.pdf,.jpeg,.png}
\else
   \usepackage[dvips]{graphicx}

   \graphicspath{{../eps/}}

   \DeclareGraphicsExtensions{.eps}
\fi
\usepackage{tablefootnote}
\usepackage[cmex10]{amsmath}
\usepackage{algorithm}
\usepackage{algorithmic}
\ifCLASSOPTIONcompsoc
\else
  \usepackage[caption=false,font=footnotesize]{subfig}
\fi
\usepackage{makecell}
\begin{document}

\title{ Multi-layer MIMO Relay as Deep Physical Neural Networks: Power Amplifiers as Activation Functions }
\author{Meng Hua,~\IEEEmembership{Senior Member,~IEEE,}
 Itsik Bergel,~\IEEEmembership{Senior Member,~IEEE}, and 
 Deniz~G\"und\"uz,~\IEEEmembership{Fellow,~IEEE}
 	\thanks{This work was supported by UKRI under the projects AI-R (EP/X030806/1) and INFORMED-AI (EP/Y028732/1), and by the SNS JU project 6G-GOALS under the EU Horizon program (Grant Agreement No. 101139232). 
}
 \thanks{M. Hua and D. G\"und\"uz are with the Department of Electrical and Electronic Engineering, Imperial College London, London SW7 2AZ, U.K. (e-mail: \{m.hua,d.gunduz\}@imperial.ac.uk). I. Bergel is with the Faculty of Engineering, Bar-Ilan University, Ramat
 	Gan 5290002, Israel (e-mail: itsik.bergel@biu.ac.il).}
}

\maketitle
\begin{abstract}
Wireless physical neural networks (WPNNs) embed neural computation directly into analog hardware, offering lower energy consumption and latency than conventional digital implementations. In this paper, we propose a deep WPNN in which nonlinear activations are realized by a multi-hop multiple-input multiple-output (MIMO) relay network, in which each relay implements a trainable complex linear gain and bias, followed by the power amplifier's intrinsic nonlinearity acting as an activation function. The cascade of multiple relays therefore realizes an over-the-air fully connected network whose parameters can be trained end-to-end. We develop two transceiver designs for different channel state information (CSI) availability scenarios: a least squares (LS)-based scheme requiring only receiver-side CSI, and a singular-value-decomposition (SVD)-based  scheme requiring both transmitter-side and receiver-side CSI. Simulation results show that the proposed architecture  enables accurate over-the-air inference for image classification. In particular, the results highlight  the advantage of exploiting hardware nonlinearity for enhanced inference capability.
\end{abstract}

\section{Introduction}
The remarkable success of artificial intelligence (AI) has been largely driven by deep learning models executed on graphics processing units (GPUs), whose   massive parallelism and  numerical precision  enable efficient training and inference. However,  GPUs inherit the von Neumann architecture, in which the physical separation between memory and computation incurs substantial energy and latency costs, particularly for large-scale  or real-time deployments. The emerging concept of the \textit{physical neural network (PNN)} offers a promising alternative \cite{wright2022deep, momeni2025training,iten2020discovering}: linear  mappings and nonlinear activations  are realized through controllable analog mechanisms, including electronic, optical, or wireless,  whose  intrinsic dynamics and nonlinearity  enable in-situ inference with significantly lower  energy and  delay.

Recently, research on implementing wireless PNNs (WPNNs) over the air has gained increasing attention \cite{hua2026WPNN}. First, physical substrates such as the phase shifts of reconfigurable intelligent surfaces (RISs) and the amplification coefficients of relays can be regarded as trainable neurons within the wireless medium. Second, fundamental algebraic operations, such as matrix multiplication and summation, can be inherently realized over the air by exploiting the superposition property of the wireless multiple-access channel \cite{Amiri2020federated}, thereby significantly reducing computation energy consumption and processing latency. Nevertheless, research on  over-the-air WPNNs  remains in its early stage, with only a limited number of studies reported to date \cite{stylianopoulos2025over,hua2025implementing,zhang2024radio, liu2022programmable, liu2025over, yangyuzhi2024realizing,hua2025aircnn, Garcia2023irNN,bergel2024nonlinear,wang2022distributed,bian2025overtheair}. In particular, \cite{stylianopoulos2025over,hua2025implementing,zhang2024radio, liu2022programmable, liu2025over, yangyuzhi2024realizing,hua2025aircnn, Garcia2023irNN} investigated the use of RISs as neural network neurons, where the adjustable phase shifts of RIS elements are treated as trainable parameters of the network. For instance, in \cite{Garcia2023irNN}, the authors proposed leveraging RISs to realize one-dimensional convolution operations by exploiting multipath propagation delays, wherein each channel impulse response acts as an individual finite impulse response filter that convolves with the transmitted signal to emulate a digital convolutional neural network. Relays have also been explored as WPNN hardware in \cite{bergel2024nonlinear,wang2022distributed,bian2025overtheair}, where the relay amplification coefficients are interpreted as analog neuron weights. When multiple relays are employed, the combined effects of wireless channels and amplification gains form an equivalent virtual multiple-input multiple-output (MIMO) system. For example, \cite{bergel2024nonlinear} demonstrated that by exploiting both the amplification gain and the inherent nonlinearity of power amplifiers (PAs), significant communication performance gains can be achieved.
However, the aforementioned studies have not fully exploited the spatial degrees of freedom or the intrinsic nonlinearity of physical devices, which fundamentally limit their expressive capacity.

\begin{figure}[!t]
	\centerline{\includegraphics[width=3.5in]{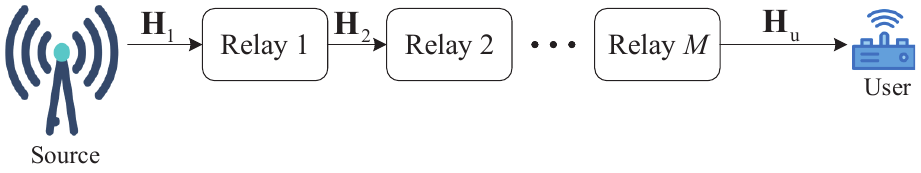}}
	\caption{The  architecture of  multi-hop MIMO relay  systems.}  \label{SystemModel}
	\vspace{-0.5cm}
\end{figure}

In this paper, we study deep physical neural networks via a multi-hop MIMO relay architecture, as illustrated in Fig.~\ref{SystemModel}.
Our main contributions are as follows. First, we propose a multi-hop MIMO relay architecture for realizing a deep WPNN, formally associating the trainable amplification, bias, and PA nonlinearity at each relay with the weight, bias, and activation function of one FC layer. Second, we propose two transceiver schemes for different channel state information (CSI) availability scenarios: a least squares (LS)-based scheme requiring only receiver-side CSI (CSIR), and a singular-value-decomposition (SVD)-based scheme requiring both transmitter-side and receiver-side CSI (CSIR/T). Both schemes support end-to-end training with standard backpropagation through the PA model. Third, through simulations on the Fashion-MNIST dataset, we show that, contrary to expectation, the SVD-based scheme that decouples eigenmodes does not benefit from PA nonlinearity in deep cascades, whereas the LS-based scheme exploits nonlinearity and scales gracefully with depth.

%

\section{System model} \label{sec:systemmodel}
As illustrated in Fig.~\ref{SystemModel}, we consider a multi-hop MIMO relay system, where information
 is transmitted from a  source to a  user through $M$ MIMO relays ${R_1,\ldots,R_M}$.  For notational convenience, we  denote the source node and the user as $R_0$ and $R_{M+1}$, respectively.  All nodes in the network are equipped with equal number of transmit and receive antennas. Specifically, let $N_m$ denote the number of transmit or receive antennas at node $R_m$, for $m \in \left\{ {0, \ldots ,M+1} \right\}$.

The objective of this relay network is to perform   a generic inference task on the signal $\bf S \in {\cal S}$
available at  $R_0$, and conveyed to $R_{M+1}$ through $M$ MIMO relays. The task is represented by $\mathcal{T} : \mathcal{S} \rightarrow \mathcal{Y}$, where $\mathcal{Y}$ denotes the task-specific output space. A concrete example of an image classification task will be introduced in Section~\ref{sec:optimziation}.   
Let ${{\mathbf{H}}_m} \in {{\mathbb C}^{{N_{m}} \times {N_{m-1}}}}$ denote the complex  baseband equivalent MIMO channel from node ${{R_{m-1}}}$ to node ${{R_{m}}}$ for $m \in \left\{ {1, \ldots ,M+1} \right\}$.  We adopt a Rician fading model for all wireless links. Specifically, the channel matrix of the $m$-th hop,  ${{\bf{H}}_m}$, is expressed as
\begin{align}
{{\mathbf{H}}_m}{\text{ = }}\sqrt {{\alpha _m}} \left( {\sqrt {\frac{K}{{K + 1}}} {{\mathbf{H}}_{m,{\text{LoS}}}} + \sqrt {\frac{1}{{K + 1}}} {{\mathbf{H}}_{m,{\text{NLoS}}}}} \right),
\end{align}
where ${{\alpha _m}}$  captures the large-scale fading of the $m$-th hop, and 
$K$ represents the Rician factor. The line-of-sight (LoS) component ${{\mathbf{H}}_{m,{\text{LoS}}}}$ is modeled as a deterministic rank-one matrix and is  given by
\begin{align}
{{\mathbf{H}}_{m,{\text{LoS}}}} = {{\mathbf{a}}_{\text{r}}}\left( {\theta _m^{\text{r}};{N_m}} \right){\mathbf{a}}_{\text{t}}^H\left( {\theta _{m - 1}^{\text{t}};{N_{m-1}}} \right),
\end{align}
where ${{\mathbf{a}}_{\text{r}}}\left( {\theta _m^{\text{r}};{N_m}} \right)$ and ${{\mathbf{a}}_{\text{t}}}\left( {\theta _{m - 1}^{\text{t}};{N_{m-1}}} \right)$ denote the receive and transmit array response vectors at nodes $R_m$ and $R_{m-1}$, respectively. The angles of arrival (AoA) and departure (AoD) associated with the LoS path are denoted by
${\theta _m^{\text{r}}}$ and ${\theta _{m - 1}^{\text{t}}}$, respectively, which are assumed to be independently and uniformly distributed over $\left[ {0,\pi } \right]$.
 For a uniform linear array with $N$ antennas, its array response is given by
\begin{align}
{{\mathbf{a}}_{\text{c}}}\left( {\theta ;N} \right) = {\left[ {1,{e^{j\pi \sin \theta }}, \ldots ,{e^{j\pi \sin \theta \left( {N - 1} \right)}}} \right]^T},{\text{c}} \in \left\{ {{\text{r,t}}} \right\}.
\end{align}
The non-LoS (NLoS) component   ${{{\mathbf{H}}_{m,{\text{NLoS}}}}}$ models the small-scale scattering, where each entry is independently distributed as  ${\left[ {{{\mathbf{H}}_{m,{\text{NLoS}}}}} \right]_{i,j}} \sim {\cal CN}\left( {0,1 } \right)$. 

Let ${{\mathbf{X}}_m} \in {{\mathbb C}^{{N_m} \times L}}$ and 
${{\mathbf{Y}}_m} \in {{\mathbb C}^{{N_m} \times L}}$ denote the transmitted  and received symbols at node $R_m$ for $m \in \left\{ {0,1, \ldots ,M + 1} \right\}$, respectively, where $L$ denotes the number of transmitted symbols, which will be specified later.  We have 
\begin{align}
{{\bf{Y}}_m} = {{\bf{H}}_m}{{\bf{X}}_{m-1}} + {{\bf{N}}_m}, ~m \in \left\{ {1, \ldots ,M + 1} \right\}, \label{signal_model_equation}
\end{align}
where ${{\mathbf{N}}_m} \in {{\mathbb C}^{{N_m} \times L}}$ denotes the additive white Gaussian noise, with each entry in ${{\bf{N}}_m}$ independently distributed as  ${\left[ {{{\bf{N}}_m}} \right]_{i,j}} \sim {\cal CN}\left( {0,{\sigma ^2}} \right)$.

We consider two cases, namely  CSIR MIMO and   CSIR/T MIMO,  depending on the availability of CSI at each node. 
\subsection{CSIR MIMO System}
In a CSIR MIMO system, the CSI is available at the receiver for each hop, which allows it to equalize the transmitted signals.  According to the input–output relationship in \eqref{signal_model_equation}, with the knowledge of ${{\mathbf{H}}_m}$,  $R_m$  can perform receiver-side signal processing to estimate the transmitted symbols. We adopt   the LS estimator  to recover  an estimate  ${{{\bf{\hat X}}}_{m - 1}}$ of ${{{\bf{X}}}_{m - 1}}$.    By applying a node-specific processing function parameterized by ${{{\bm{\theta }}_m}}$, the transmitted signal at 
$R_m$ can be expressed as
\begin{align}
	{{\bf{X}}_m} = {g_{{{\bm{\theta }}_m}}}\left( {{{{\bf{ Y}}}_{m }},{{\bf{H}}_m},{\sigma ^2}} \right), ~m \in \left\{ {1, \ldots ,M + 1} \right\}, 
\end{align}
where ${g_{{{\bm{\theta }}_m}}}\left(  \cdot  \right)$ denotes the signal processing function implemented at  $R_m$. 
\subsection{CSIR/T MIMO System}
In a  CSIR/T MIMO system, the CSI is available at both the transmitter and receiver for each hop. Therefore,  the encoding function parameterized by WPNN parameters ${{{\bm{\phi }}_m}}$ at   $R_m$  can be expressed as 
\begin{align}
\!\!\!{{\bf{X}}_m} = {f_{{{\bm{\phi }}_m}}}\left( {{{{\bf{ Y}}}_{m}},{{\bf{H}}_m},{{\bf{H}}_{m + 1}},{\sigma ^2}} \right), ~m \in \left\{ {1, \ldots ,M } \right\},
\end{align}
where ${f_{{{\bm{\phi }}_m}}}\left(  \cdot  \right)$ denotes the signal processing function implemented at $R_m$. 

The WPNN realized by the multi-hop MIMO relay network is trained end-to-end to approximate the task mapping $\mathcal{T}$. Denoting by $\widehat{\mathcal{T}}_{\bm \Theta}({\bf S})$ the mapping from the input sample $\bf S$ to the user's decision, parameterized by all trainable physical-layer parameters $\bm \Theta$ (amplification matrices, biases, transmit/receive processing, and final read-out layer), the objective is to minimize a task-specific loss: $\min_{\bm \Theta} \mathbb{E}_{({\bf S},y) \sim P_{{\bf S},y}} [{\cal L}(\widehat{\mathcal{T}}_{\bm \Theta}(S), y)]$, where $y \in \mathcal{Y}$ is the ground-truth label.

\section{Role of PA Nonlinearity in Multi-Hop MIMO Relay Systems}
To motivate the role of PA nonlinearity, we contrast linear and nonlinear PA regimes.

$\textbf{Linear PA}$: Assume that the PA at each relay  operates strictly in its linear region so that ${{\mathbf{X}}_m} = {{\mathbf{W}}_m}{{\mathbf{Y}}_m}$, where ${{\mathbf{W}}_m}\in {{\mathbb C}^{{N_m} \times N_m}}$  is the linear amplification matrix.  Substituting it into  \eqref{signal_model_equation} recursively yields 
\begin{align}
{{\mathbf{Y}}_{M + 1}} = {{\mathbf{H}}_{M + 1}}{{\mathbf{W}}_M}{{\mathbf{H}}_M}{{\mathbf{W}}_{M - 1}} \cdots {{\mathbf{H}}_1}{{\mathbf{X}}_0} + {{\mathbf{N}}_{{\text{eff}}}}, \label{user_received_signal}
\end{align}
where ${{\mathbf{N}}_{{\text{eff}}}}$ aggregates  the per-hop noise and is independent of  ${\bf X}_0$. The end-to-end map  ${{\mathbf{X}}_0} \to {{\mathbf{Y}}_{M + 1}}$ is therefore linear and collapses to a single composite FC layer for any $M$, fundamentally limiting the expressive capability of the WPNN. Basically, the achievable function class is the set of complex linear maps of limited rank $\mathop {\min }\limits_m {\text{rank}}\left( {{{\mathbf{H}}_m}{{\mathbf{W}}_m}} \right)$.

$\textbf{Nonlinear PA}$:
When the PA  exhibits a  nonlinear transfer characteristic, the relay output becomes 
\begin{align}
{{\mathbf{X}}_m} = {\text{PA}}\left( {{{\mathbf{W}}_m}{{\mathbf{Y}}_m}} \right),
\end{align}
where ${\rm{PA}}\left(  \cdot  \right)$  denotes  the  element-wise nonlinear input–output map,  whose specific form will be detailed in  Section \ref{sec:optimziation}. Each relay hop realizes linear mixing followed by a pointwise nonlinear feature transformation, so the  
$M$-hop relay chain is equivalent to an $M$-layer FC network with implicit activations. This breaks the limited rank bound of linear PA, making the achievable function class grow with $M$. This is the structural basis  for interpreting the multi-hop MIMO relay system as  a deep WPNN.

\section{Deep Learning Optimization  with Nonlinear PA} \label{sec:optimziation}
In this section, we instantiate the general framework developed in Section \ref{sec:systemmodel} for an image classification task, while noting that the proposed design can be readily extended to other inference tasks. We then present the training strategy and the corresponding loss function used to train the WPNN for this task.
\subsection{Transmission Architecture Design} \label{subsection: architecture_design}
\subsubsection{CSIR Design}
 Let the input sample be an image ${\bf S} \in {{\mathbb R}^{C \times H \times W}}$, where $C$, $H$, and $W$ represent the number of color channels, height, and width, respectively, and the output space $\mathcal {Y}$ is the discrete set of image classes.
The source node $R_0$ first normalizes  $\bf S$ so that its pixel values lie in $\left[ {0,1} \right]$. The normalized image is then vectorized and reshaped  into a complex-valued matrix 
  ${{\bf{S}}_c} \in {{\mathbb C}^{{N_{ 0}} \times L}}$ with $L = \frac{{C \times H \times W}}{{2{N_{0}}}}$. Then, the  signal transmitted by $R_0$ is given by
\begin{align}
{{\mathbf{X}}_0} = {\text{PA}}\left( {{{\mathbf{F}}_0}{{\mathbf{S}}_c} + {{\mathbf{b}}_0}{\mathbf{1}}_L^T} \right),
\end{align}
where  ${{\bf{F}}_0} \in {{\mathbb C}^{{N_{0}} \times {N_{0}}}}$ and ${{\bf{b}}_0} \in {{\mathbb C}^{{N_{0}} \times 1}}$  denote the  precoder matrix  and the bias vector, respectively, and ${{\mathbf{1}}_L}$ is the $L$-length vector of all ones. 
The bias vector ${{\bf{b}}_0}$ plays a role analogous to that in digital neural networks and can be practically realized by injecting a direct current  offset, which is trainable. The  Rapp PA model, denoted by ${\rm{PA}}\left(  \cdot  \right)$,   can be modeled as  \cite{wang2022distributed}
\begin{align}
{\text{PA}}\left( x \right) = \frac{x}{{{{\left( {1 + {{\left( {{{\left| x \right|} \mathord{\left/
										{\vphantom {{\left| x \right|} {{x_{{\text{sat}}}}}}} \right.
										\kern-\nulldelimiterspace} {{x_{{\text{sat}}}}}}} \right)}^{2p}}} \right)}^{{1 \mathord{\left/
						{\vphantom {1 {\left( {2p} \right)}}} \right.
						\kern-\nulldelimiterspace} {\left( {2p} \right)}}}}}},
\end{align}
where $p$ and $x_{\rm sat}$ represent the PA parameters. Following \cite{bergel2024nonlinear}, we set  $p=2$ and $x_{\rm sat}=1$, under which the amplitude response of the nonlinear power amplifier exhibits a smooth saturation behavior similar to that of a $\mathrm{tanh}$ function.
Accordingly, the signal received  at relay  $R_1$ can be represented as 
\begin{align}
	{{\bf{Y}}_1} = {{\bf{H}}_1}{{\bf{X}}_0} + {{\bf{N}}_1}.
\end{align}
Then, a LS MIMO  estimator  is employed to exploit the CSI to decouple the entangled signal ${{\bf{Y}}_1}$ as ${{{\bf{\hat X}}}_0}$:
\begin{align}
{{{\bf{\hat X}}}_0} = {\bf{H}}_1^{+} {{\bf{Y}}_1} = {\bf{H}}_1^{+} \left( {{{\bf{H}}_1}{{\bf{X}}_0} + {{\bf{N}}_1}} \right), \label{souce_R0}
\end{align} 
where ${\left(  \cdot  \right)^ + }$ denotes the  the Moore–Penrose pseudo-inverse.
Next, a trainable relay amplification  matrix, denoted by ${{\bf{F}}_1} \in {{\mathbb C}^{{N_1} \times N_0}}$, is applied to scale ${{{\bf{\hat X}}}_0} $, and  a trainable bias vector ${{\bf{b}}_1} \in {{\mathbb C}^{{N_1} \times 1}}$ is added,   and the result passes through the nonlinear PA. The  output signal at relay $R_1$ can thus be expressed as
\begin{align}
{{\mathbf{X}}_1} = {\text{PA}}\left( {{{\mathbf{F}}_1}{{{\mathbf{\hat X}}}_0} + {{\mathbf{b}}_1}{\mathbf{1}}_L^T} \right).
\end{align}
It can be seen that this nonlinear transformation plays the role of an activation function, enabling the relay to realize both amplification and nonlinear feature mapping over the air. Therefore, the relay effectively performs one FC layer, where the amplification matrix ${{{\bf{F}}_1}}$ and the bias vector ${{{\bf{b}}_1}}$ correspond to the trainable weights and bias of a conventional digital neural network, respectively.

At the subsequent relay nodes ${R_2}, \ldots ,{R_M}$, similar signal processing operations are performed, where each relay employs its own trainable amplification matrix and bias vector, followed by the inherent nonlinear amplifier characteristic. Consequently, the entire multi-hop relay chain can be viewed as a cascade of  over-the-air FC layers, thereby forming a multi-layer WPNN. In this sense, the CSIR MIMO relay network inherently realizes a deep neural architecture in the analog domain, with each relay acting as one neural layer that jointly contributes to the end-to-end inference process.
\subsubsection{CSIR/T Design}
Given ${{\mathbf{H}}_m} \in {{\mathbb C}^{{N_{m}} \times {N_{m-1}}}}$, 
we first decompose the channel matrix ${{\bf{H}}_m}$ by SVD, yielding ${{\bf{H}}_m} = {{\bf{U}}_m}{{\bf{\Sigma }}_m}{\bf{V}}_m^H$, where ${{\mathbf{U}}_m} \in {{\mathbb C}^{{N_m} \times {N_m}}}$ and ${{\mathbf{V}}_m} \in {{\mathbb C}^{{N_{m - 1}} \times {N_{m - 1}}}}$ are unitary matrices, and  ${{\bf{\Sigma }}_m} \in {{\mathbb C}^{{N_{m}} \times {N_{m - 1}}}}$ is a diagonal matrix whose
singular values are real and sorted in a descending order. For a CSIR/T MIMO system, the CSI can be  leveraged at the transmitter side, and the output signal at source $R_0$ is given by 
\begin{align}
{{\mathbf{X}}_0} = {\text{PA}}\left( {{{\mathbf{V}}_1}{{\mathbf{F}}_0}{{\mathbf{S}}_c} + {{\mathbf{b}}_0}{\mathbf{1}}_L^T} \right).
\end{align}
At relay $R_1$, the received signal is first processed by the combiner ${\bf{U}}_1^H$ to decouple the spatial streams according to the SVD structure of ${{{\bf{H}}_1}}$.  The resulting signal is then scaled by the pseudo-inverse of the singular-value matrix  ${\bf{\Sigma }}_1^{+} $ to normalize the power across the eigenmodes. Subsequently, an amplification  matrix  ${{{\bf{F}}_1}}$  
is applied to control the relay gain and adapt the transmitted power level. Finally, the processed signal is multiplied by the right singular matrix ${{{\bf{V}}_2}}$, which serves as a pre-processing operation aligned with the channel ${{{\bf{H}}_2}}$ toward the next hop. This process at $R_1$ can be written as 
\begin{align}
{{\mathbf{X}}_1} = {\text{PA}}\left( {{{\mathbf{V}}_2}{{\mathbf{F}}_1}{\mathbf{\Sigma }}_1^ + {\mathbf{U}}_1^H{{\mathbf{Y}}_1} + {{\mathbf{b}}_1}{\mathbf{1}}_L^T} \right).
\end{align}
This sequential combination of receive combining, singular-value equalization, amplification, and transmit pre-processing can be extended to subsequent relays, yielding an end-to-end mapping whose parameters $\left\{ {{{\mathbf{F}}_m},{{\mathbf{b}}_m}} \right\}_{m = 1}^M$ 
are jointly trained.
\subsection{Communication Design}
Based on Subsection \ref{subsection: architecture_design}, the  signal received at the user after signal processing is  given by 
\begin{align}
{{\mathbf{X}}_{M + 1}} = \left\{ {\begin{array}{*{20}{l}}
	{{{\mathbf{F}}_{M + 1}}{{{\mathbf{\hat X}}}_M} + {{\mathbf{b}}_{M + 1}}{\mathbf{1}}_L^T,{\kern 1pt} {\kern 1pt} {\kern 1pt} {\kern 1pt} {\kern 1pt}  {\kern 1pt}  {\kern 1pt}  {\kern 1pt} {\kern 1pt}{\kern 1pt}{\kern 1pt}{\kern 1pt}{\kern 1pt}{\kern 1pt}{\kern 1pt}  {\kern 1pt} \qquad \qquad \qquad  {\text{CSIR}},} \\ 
	{{{\mathbf{F}}_{M + 1}}{\mathbf{\Sigma }}_{M + 1}^ + {\mathbf{U}}_{M + 1}^H{{\mathbf{Y}}_{M + 1}} + {{\mathbf{b}}_{M + 1}}{\mathbf{1}}_L^T,{\kern 1pt} {\kern 1pt} {\kern 1pt} {\text{CSIR/T}},} 
	\end{array}} \right.
\end{align}
where ${{\mathbf{F}}_{M+1}}\in {{\mathbb C}^{{N_{M+1}} \times {N_{M}}}}$ and  ${{\bf{b}}_{M+1}} \in {{\mathbb C}^{{N_{M+1}} \times 1}}$ denote the combiner and bias at user, respectively, and  
 ${{{\bf{\hat X}}}_M}$ denotes the estimated signals transmitted from $R_M$, which is similarly defined in \eqref{souce_R0}.
After obtaining the processed signal ${{\bf{X}}_{M+1}}$, it is first converted into its real-valued representation by concatenating the real and imaginary parts. The resulting real-valued tensor is then flattened into a one-dimensional feature vector, which is fed into a task-specific read-out layer to produce the final inference output. For the image classification example, the read-out layer consists of an FC layer followed by a softmax activation  producing the class probability vector ${{\mathbf{\hat p}}}$ over the $C$ classes.

\subsection{Training Loss}
For the image classification task, a standard cross-entropy loss is adopted to train the WPNN, expressed as
\begin{align}
{{\cal L}_{{\rm{loss}}}} =  - \sum\limits_{i = 1}^C {{p_i}\log \left( {{{\hat p}_i}} \right)},  \label{lossfunction}
\end{align}
where $C$ denotes the number of classes, ${{p_i}}$ is the one-hot true label, and ${{{\hat p}_i}}$ is the predicted  probability of the $i$th class. For other tasks, ${{\cal L}_{{\rm{loss}}}}$ would be replaced by the corresponding task-specific loss function without changing the underlying architecture or the training procedure.

\section{Numerical results} \label{sec:numerial_results}
We evaluate the proposed multi-hop MIMO relay-based WPNN in terms of classification accuracy on the Fashion-MNIST dataset, which contains 60,000 training examples and 10,000 test examples across 10 categories, where each example is a $28 \times 28$ grayscale image.
The path loss between the source node and the user is normalized to one. 
 The relay nodes are uniformly placed along the line connecting the source node and the user, yielding ${\alpha _m} = {\left( {M + 1} \right)^{2}}$. 
The signal-to-noise ratio (SNR) is defined as ${\rm{SNR = 10lo}}{{\rm{g}}_{10}}\frac{{{1}}}{{{\sigma ^2}}}$ in dB. 
Moreover, the average transmit power at the linear PA is set to  $1$ W. Unless specified otherwise, we set  $N_0=28$, $L=14$,  and $N_m=32$, $m \in \left\{ {1, \ldots ,M + 1} \right\}$.   The proposed WPNN is trained in an end-to-end manner using the Adam optimizer with a learning rate of $10^{-4}$ and a batch size of 64. During training, one independent channel realization is generated for each image transmission. 

The following  schemes are considered:
\begin{itemize}
	\item \textbf{Upper bound:} This scheme employs an ideal digital neural network with the same layer dimensions and depth as the proposed WPNN. Each relay-associated physical layer is replaced by a perfect digital FC layer, without wireless-channel distortion. The nonlinear activation is implemented by the standard $\tanh(\cdot)$ function.
	\item \textbf{LS, LPA:} This scheme corresponds to the CSIR case, requiring only   receiver CSI for  each hop. All PAs are modeled as ideal linear amplifiers.
	\item \textbf{LS, \!\!NPA:} This scheme adopts the same LS-based transceiver architecture as ``LS, LPA'', except that the nonlinear Rapp PA model is employed.
	\item \textbf{SVD, LPA:} This scheme corresponds to the CSIR/T case, exploiting both transmitter- and receiver-side CSI for each hop. All PAs are modeled as ideal linear amplifiers.
	\item \textbf{SVD, NPA:} This scheme adopts the same SVD-based transceiver architecture as ``SVD, LPA'', except that the nonlinear Rapp PA model is employed.
	\item \textbf{Training-testing PA mismatch (TPM):} The WPNN is trained assuming ideal linear PAs but is evaluated using nonlinear PAs, without retraining or fine-tuning. This mismatch setting is considered for both the LS- and SVD-based transceiver architectures.


\end{itemize}

\begin{figure}[!t]
	\centerline{\includegraphics[width=3.2in]{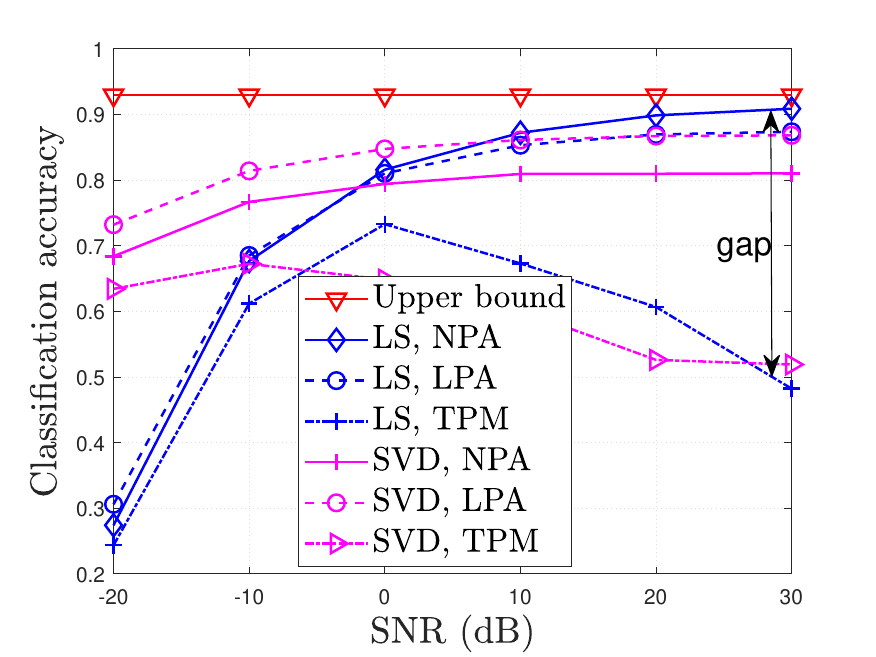}}
	\caption{Classification accuracy versus SNR.}  \label{SNR_vs_CA}
	\vspace{-0.5cm}
\end{figure}

Fig.~\ref{SNR_vs_CA} compares the classification accuracy of different transceiver schemes under varying SNRs for   $M=1$ relay and $K =  - \infty $ (dB). 
We can observe that for SNRs above $0$~dB, the LS scheme with nonlinear PA outperforms  linear PA due to its superior expressiveness. In particular, at an SNR of $20$~dB, the performance of ``LS, NPA'' scheme is already close to the upper bound.  For the SVD-based scheme, the linear PA consistently performs better since the nonlinearity prevents full eigenmode decoupling, leaving residual inter-stream interference.  
Moreover, the TPM schemes under both the LS- and SVD-based architectures exhibit clear performance losses relative to their nonlinear-PA counterparts. These losses become more pronounced as the SNR increases, since the inconsistency between the assumed and actual PA models becomes the dominant performance-limiting factor when noise is weak.
These results illustrate that by appropriately tuning the hardware parameters at each node, the proposed WPNN can closely approximate the performance of a digital neural network.
\begin{figure}[!t]
	\centerline{\includegraphics[width=3.2in]{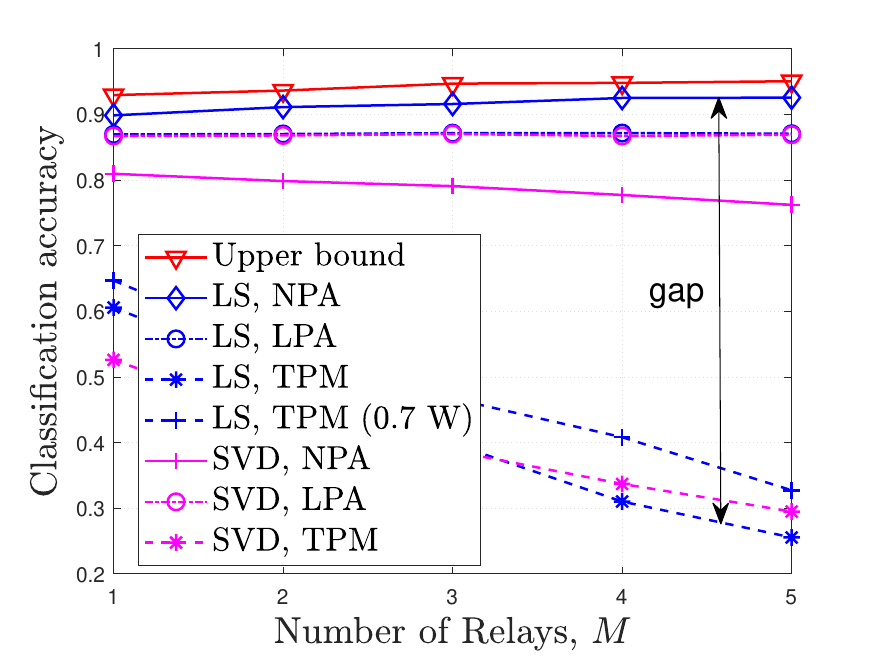}}
	\caption{Classification accuracy   versus number of relays $M$.}  \label{Relay_vs_CA}
	\vspace{-0.5cm}
\end{figure}

Fig.~\ref{Relay_vs_CA} illustrates the impact of the number of relay nodes $M$ on the classification accuracy for different transceiver schemes under ${\rm SNR}=20$~dB and $K =  - \infty $ (dB). 
For the “LS, NPA’’ scheme, the classification accuracy  increases with $M$, from  0.8988 at $M=1$ to 0.9258 at $M=5$, and  consistently outperforms its linear counterpart across all $M$.  
 In contrast, 
 the “SVD, NPA’’ scheme exhibits noticeable degradation as $M$ increases, since the nonlinear PA distorts the SVD-based beamforming structure and such distortion accumulates across multiple hops. Also, for both linear PA schemes,  the accuracy remains  the same with increasing 
$M$ due to the absence of nonlinear activation, which limits the expressive capability of the cascaded architecture.
It is further observed that the accuracy of TPM schemes under both the LS- and SVD-based architectures diminishes with  $M$. This is due to the accumulation of errors caused by PA-model mismatch over multiple hops. We further consider two transmit-power constraints for the TPM scheme. Specifically, the ``LS, TPM (0.7 W)” scheme corresponds to the case in which the relays are trained with an average transmit power of 0.7 W. Interestingly, ``LS, TPM (0.7 W)” outperforms ``LS, TPM” because the lower training power keeps the PAs closer to their linear operating region, thereby reducing the mismatch between the linear PA model assumed during training and the nonlinear PA behavior encountered during testing.

\begin{figure}[!t]
	\centerline{\includegraphics[width=3.2in]{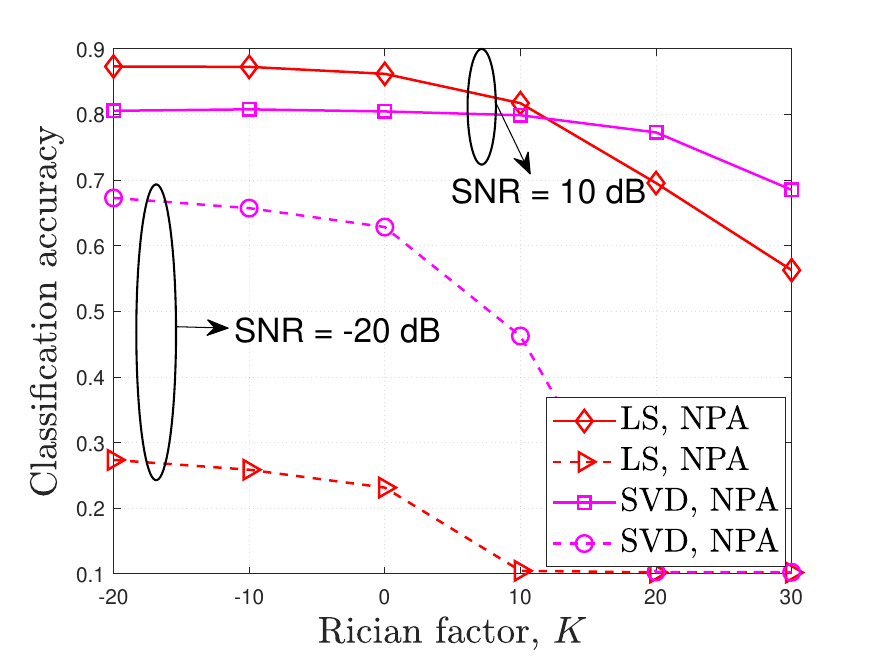}}
	\caption{Classification accuracy versus Rician factor $K$.}  \label{fig:vs_Ricianfactor}
	\vspace{-0.5cm}
\end{figure}

Fig.~\ref{fig:vs_Ricianfactor}  illustrates the impact of the Rician factor $K$ on the classification accuracy for $M=1$ under $\mathrm{SNR}=-20~\mathrm{dB}$ and $\mathrm{SNR}=10~\mathrm{dB}$. It can be  observed that the classification accuracy generally degrades as $K$ increases for both LS- and SVD-based schemes. 
This is because lower-rank channels destroy the spatial
degrees-of-freedom on which the WPNN relies.
At $\mathrm{SNR}=10~\mathrm{dB}$, the accuracy still decreases with $K$, but the drop is much less drastic since the noise is no longer the dominating impairment during per-hop equalization. For instance, even at $K=20~\mathrm{dB}$, the LS- and SVD-based schemes achieve accuracies of $0.6957$ and $0.7728$, respectively.

\section{Conclusion}
This paper has proposed a deep WPNN realized through a multi-hop MIMO relay network, in which each relay implements a trainable linear precoding stage followed by the intrinsic nonlinear activation of its PA. Cascading $M$ such relays yields an over-the-air multi-layer FC network that unifies communication and computation over the same wireless infrastructure.  Two transceiver designs were developed:  an LS-based scheme requiring only receiver CSI, and an SVD-based scheme exploiting joint transmitter–receiver CSI.  Three main findings emerged from our study: First, PA nonlinearity is a resource, rather than merely an impairment,  for  over-the-air computing. The LS scheme with a nonlinear PA monotonically improves with $M$. Second, this improvement is architecture-dependent: nonlinearity disrupts SVD eigenmode decoupling, so a linear-PA design is preferable for CSIR/T systems. Third, hardware-model mismatch and increasing channel rank deficiency (large Rician $K$) cause errors that compound over hops, motivating mismatch-aware training. Two natural extensions are: (i) more complex over-the-air architectures beyond FC layers, e.g., convolutional or attention-based WPNNs; and (ii) robust, mismatch- and CSI-uncertainty-aware end-to-end training for deployment under realistic hardware imperfections.

\bibliographystyle{IEEEtran}
\bibliography{MIMO_Relay_PNN}
\end{document}